\pdfoutput=1

\documentclass[11pt,a4paper]{article}

\usepackage[hyperref]{conll2018}
\usepackage{times}
\usepackage{latexsym}

\usepackage{url}

\aclfinalcopy 

\usepackage[shortlabels,inline]{enumitem}
\usepackage{graphicx}
\usepackage{subfigure} 
\usepackage{multirow}
\usepackage{amsmath}
\usepackage{amsfonts}
\usepackage{color}
\usepackage{xcolor}
\usepackage{tikz}
\usepackage{pgfplots} 
\usepgfplotslibrary{colorbrewer}
\pgfplotsset{compat=1.9}

\definecolor{bblue}{HTML}{4F81BD}
\definecolor{rred}{HTML}{C0504D}
\definecolor{ggreen}{HTML}{9BBB59}
\definecolor{ppurple}{HTML}{9F4C7C}

\usepackage{CJKutf8}

\usepackage{makecell}

\title{Exploiting Effective Representations for Chinese Sentiment Analysis Using a Multi-Channel Convolutional Neural Network}

\author{Pengfei Liu, Ji Zhang, Cane Wing-Ki Leung, Chao He and Thomas L. Griffiths \\ \\
  Wisers AI Lab, Wisers Information Limited, Hong Kong \\
  {\tt \small{ppfliu@gmail.com, \{jasonzhang,caneleung,chaohe\}@wisers.com, tom\_griffiths@berkeley.edu}}}
\date{}

\begin{document}
\maketitle
\begin{abstract}

    Effective representation of a text is critical for various natural language processing tasks. For the particular task of Chinese sentiment analysis, it is important to understand and choose an effective representation of a text from different forms of Chinese representations such as \textit{word}, \textit{character} and \textit{pinyin}.
    This paper presents a systematic study of the effect of these representations for Chinese sentiment analysis by proposing a multi-channel convolutional neural network (MCCNN), where each channel corresponds to a representation.
    Experimental results show that: (1) Word wins on the dataset of low OOV rate while character wins otherwise; (2) Using these representations in combination generally improves the performance; (3) The representations based on MCCNN outperform conventional ngram features using SVM; (4) The proposed MCCNN model achieves the competitive performance against the state-of-the-art model \texttt{fastText} for Chinese sentiment analysis.

\end{abstract}

\section{Introduction}


Research in Chinese sentiment analysis is still substantially underdeveloped \cite{peng2017review}, although the task of sentiment analysis has been actively studied in the past decades. Conventional manual feature engineering methods are still popular in Chinese sentiment analysis, e.g., the study of lexicon-based and machine learning-based approaches for sentiment classification of Chinese microblogs by \cite{yuan2013sentiment}, the application of the maximum entropy technique for Chinese sentiment analysis by \cite{lee2011chinese}. This kind of methods require a number of feature extraction steps such as performing Chinese word segmentation, applying conjunction rules, removing stop words and punctuation, combining a negation word with the following word and counting the number of positive or negative keywords and so on.
Nonetheless, various deep learning models have been recently proposed for English sentiment analysis, such as convolutional neural networks (CNNs), \cite{kim2014convolutional,johnson2015effective}, recurrent neural networks (RNNs), \cite{tang2015document,xu2016cached}, memory networks \cite{dou2017capturing}, attention models \cite{yang2016hierarchical,zhou2016attention} and so on. These models typically learn features automatically from raw English text, yet outperform conventional methods based on manual feature engineering.


Due to linguistic differences between Chinese and English such as the lack of word delimiters in Chinese and the use of alphabets in English, deep learning models for English are not directly applicable to Chinese. To apply these models, the most common approach is to perform Chinese word segmentation and treat each Chinese word the same as an English word, named as the \textit{word representation} of Chinese. However, word representation typically leads to a very large vocabulary and different Chinese word segmentation standards (e.g., CTB, PKU, MSR) also affect the model performance \cite{pan2016chinese}. Another approach to handle Chinese is called \textit{character representation}, which treats each individual Chinese character as a single unit and thereby leads to a smaller vocabulary of Chinese characters. The character representation of Chinese has been adopted for various NLP tasks, e.g., Chinese word segmentation \cite{liu2014learning} and Chinese named entity recognition \cite{dong2016character}. The third kind of Chinese representation is called \textit{pinyin}, which is a phonetic romanization of Chinese. The pinyin representation is directly applicable in sophisticated deep learning models based on English characters \cite{zhang2015character}, and also has the advantage of leading to the smallest vocabulary since many Chinese characters share the same pinyins.

Although \textit{word}, \textit{character} and \textit{pinyin} representations of Chinese have been successfully adopted in various applications, there is still a lack of a systematic study of these representations for Chinese sentiment analysis. Such a study is practically very useful for developing a high performance sentiment analysis system. This work is the first systematic study of the effect of the three representations and their combinations for Chinese sentiment analysis, for which we propose a model called the multi-channel convolutional neural network (MCCNN) to integrate the three representations in the same model. 
We adopt a CNN as our base model because CNNs have been successfully applied in various text classification tasks due to their superior performance in extracting effective local features from a sequence of text. We choose CNNs instead of RNNs since it is reported that CNNs outperform RNNs in several sentiment analysis tasks \cite{kim2014convolutional,zhang2015character}, and RNNs are  computationally more demanding than CNNs when processing large-scale datasets.

The contributions of this paper are two-fold:
\begin{enumerate}[(1)] \setlength{\itemsep}{0pt}
    \item Presents a systematic study of the effect of different Chinese representations and their combinations for Chinese sentiment analysis;
    \item The proposed MCCNN model achieves competitive performance against the state-of-the-art \texttt{fastText} model for Chinese sentiment analysis.
\end{enumerate}

\section{Related Work}

\begin{description}[style=unboxed,leftmargin=0cm]
    
    \item[Chinese Representation]
    
    A Chinese text can be transformed into various representation forms such as words, characters, pinyins or character radicals. Many studies have been proposed to exploit these representations for various Chinese natural language processing tasks.
    For example, \newcite{zhang2015character} transformed a Chinese text into pinyin form to adopt the same character-level CNN model as English for text classification tasks.
    \newcite{chen2015joint} proposed learning word and character embeddings jointly by a character-enhanced word embedding model.
    \newcite{shi2015radical} proposed radical embedding to process Chinese characters at a deeper radical level.
    Similarly, \newcite{yin2016multi} proposed the multi-granularity embedding model to learn Chinese embeddings for words, characters and radicals jointly. 
    Further, \newcite{yu2017joint} covered more subcharacter components of Chinese characters besides radicals by an extended CBOW model \cite{mikolov2013efficient} to learn embeddings for words, characters and subcharacter components jointly.
    However, these Chinese representations are studied on different downstream tasks, making it difficult to compare their effectiveness.

    \item[Deep Learning for Sentiment Analysis]
    
     Sentiment analysis or opinion mining aims to study people's sentiments or opinions towards entities such as products, services, organizations and their attributes \cite{zhang2018deep}. Deep learning models such as CNNs and RNNs have been widely studied for text classification, including sentiment analysis, topic classification and news categorization \cite{kim2014convolutional,johnson2014effective,irsoy2014opinion,zhang2015character,conneau2016very,johnson2016convolutional}.
    For example, \newcite{dos2014deep} presented a deep CNN exploiting character- and sentence-level information for sentiment analysis of short texts.
    \newcite{conneau2016very} proposed a character-level very deep CNN (VDCNN, up to 29 convolutional layers) for text classification and showed that depth improved classification performance. In contrast, \newcite{johnson2016convolutional} reported the performance of a shallow word-level CNN, which outperforms VDCNN on the same dataset and runs much faster. Similarly, \newcite{joulin2016bag} presented the model named \texttt{fastText}, which is a shallow 3-layer neural network for general-purpose text classification tasks and obtains comparable or even better performance than the recently proposed deep models.
    For Chinese sentiment analysis, \newcite{hao2017improving} introduced a segmentation-based parallel CNN model, which first segments a text into smaller units based on punctuation marks and then feeds the units to a parallel CNN simultaneously; \newcite{wang2017chinese} presents a bilinear character-word CNN model which represents a Chinese text as a bilinear combination of CNN features obtained from both character-level and word-level embeddings. 

\end{description}

Different from the aforementioned studies, our work in this paper particularly aims to exploit effective representations for Chinese sentiment analysis and conduct a systematic study on these representations using a multi-channel CNN model where each channel corresponds to a specific Chinese representation.

\section{Approach}

\begin{figure}[htb]
	\centering
    \includegraphics[width=0.9\linewidth]{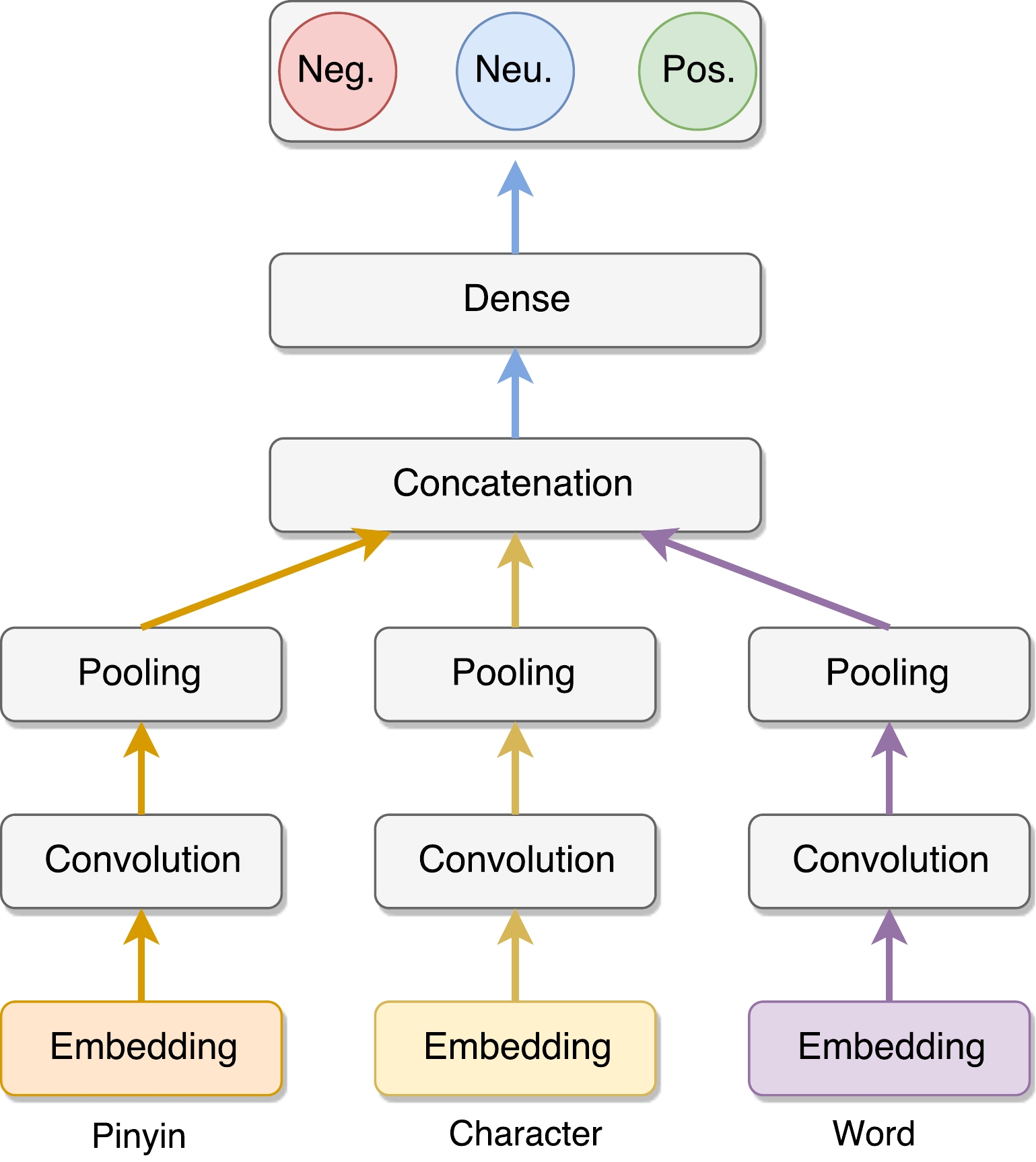} 
	\caption{A multi-channel convolutional neural network (MCCNN) for Chinese sentiment analysis.}
	\label{fig:mcnn}
\end{figure}

\begin{figure*}[htb]
	\centering
    \includegraphics[width=0.76\linewidth]{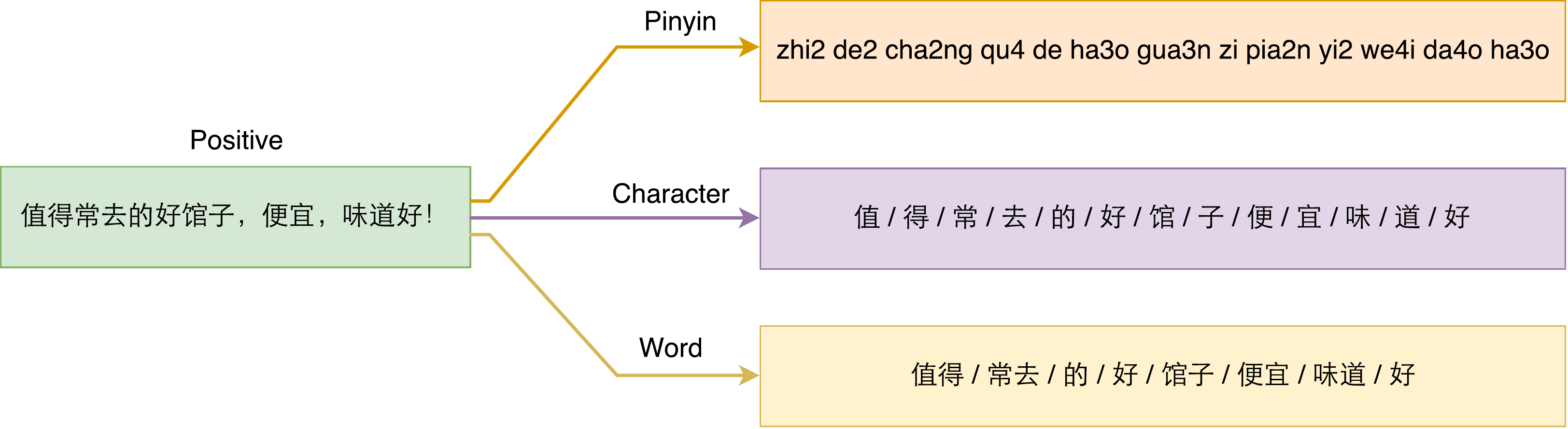} 
	\caption{A Chinese sentence in pinyin, character and word representations.}
	\label{fig:channels}
\end{figure*}

We propose a multi-channel convolutional neural network (MCCNN) to study the effect of different Chinese representations and their combinations in Chinese sentiment analysis. Figure~\ref{fig:mcnn} shows the MCCNN model with the three channels corresponding to the three representations of a Chinese text, namely \textit{pinyin}, \textit{character} and \textit{word} respectively.
Each channel in MCCNN consists of one \texttt{embedding} layer, one \texttt{convolution} layer, and one \texttt{pooling} layer. A Chinese text is transformed into the three representations and fed into the corresponding channels, whose activations are concatenated to feed a fully-connected \texttt{dense} layer. The final output layer is activated by \texttt{softmax} to predict the posterior probability of each label (i.e., \textit{positive}, \textit{neutral} and \textit{negative}).
Figure~\ref{fig:channels} shows an example of a Chinese sentence in the pinyin, character and word representations.

\section{Experiments}

\subsection{Datasets}

We conducted experiments based on four datasets as shown in Table~\ref{tbl:dataset}, which presents for each dataset the number of instances and their average number of tokens (\textit{Length}), as well as the out-of-vocabulary rate (\textit{OOVR}) of the test set under the word, character and pinyin representations.
The first two datasets are compiled by ourselves, labeled with 3 classes: \textit{positive}, \textit{negative} and \textit{neutral}. The \textit{Customer-Reviews} (CR) dataset consists of reviews from 14 industries, where the top 5 industries with the largest number of reviews are government, skincare, automobile and insurance. 
The \textit{Amazon-Dianping} (AD) dataset contains customer reviews from Amazon (\url{www.amazon.cn}) and Dianping (\url{www.dianping.com}). It covers a wide range of domains, where the top 3 domains are restaurant, shopping and domestic services. Each review comes with a rating from 1 to 5 stars. Reviews with ``1-2" stars are labeled as \textit{negative} while those with ``4-5" stars are labeled as \textit{positive}. The remaining reviews with 3 stars are labeled as \textit{neutral}.

The \textit{PKU-Product-Reviews} (Product) dataset, introduced by \newcite{wan2011bilingual}, consists of IT product reviews crawled from IT168 (\url{www.it168.com}), while the \textit{ChnSentiCorp-4000} (Hotel) dataset, firstly presented in \cite{zagibalov2008automatic}, contains hotel reviews from Ctrip (\url{www.ctrip.com}). The two datasets are labeled into 2 classes: \textit{positive} and \textit{negative}. Since they are publicly available and widely used in Chinese sentiment analysis research, they are used to evaluate the MCCNN model with different Chinese representations, and conduct performance comparison with \cite{zhai2011exploiting}, which exploited effective features for Chinese sentiment analysis based on the two datasets.

\begin{table*}[htb]
	\centering
	\caption{Dataset statistics.}
	\label{tbl:dataset}
	\resizebox{\linewidth}{!}{%
\begin{tabular}{c|c|c|c|c||c|c|c|c|c|c}
\hline
\multirow{2}{*}{\textbf{\textbf{Dataset}}} & \multirow{2}{*}{\textbf{\textbf{\#Positive}}} & \multirow{2}{*}{\textbf{\textbf{\#Negative}}} & \multirow{2}{*}{\textbf{\#Neutral}} & \multirow{2}{*}{\textbf{Total}} & \multicolumn{2}{c|}{\textbf{Word}} & \multicolumn{2}{c|}{\textbf{Character}} & \multicolumn{2}{c}{\textbf{Pinyin}} \\ \cline{6-11} 
 &  &  &  &  & $ \textbf{Length} $ & \textbf{OOVR} (\%) & \textbf{Length} & \textbf{OOVR(\%)} & \textbf{Length} & \textbf{OOVR(\%)} \\ \hline \hline
CR & 44296 & 36201 & 18120 & 98617 & 141 & 14.48 & 250 & 1.36 & 250 & 0.15 \\
AD & 80005 & 79999 & 79996 & 240000 & 45 & 11.73 & 73 & 1.06 & 73 & 0.14 \\ \hline
Product & 451 & 435 & 0 & 886 & 34 & 40.19 & 57 & 12.96 & 57 & 6.16 \\
Hotel & 2000 & 2000 & 0 & 4000 & 69 & 28.50 & 116 & 7.21 & 116 & 2.26 \\ \hline
\end{tabular}%
}
\end{table*}

\subsection{Metrics}

The average $F_1$ weighted by the number of instances within each class, and the \textit{accuracy} metric are used for experimental evaluation.
The paired \textit{t}-test is adopted to determine if two sets of results are significantly different from each other at a significance level of $p$.

\subsection{Settings}

We conduct 10-fold cross validation on the first two datasets (CR and AD) and report average results over the 10 folds.
For fair comparison with \newcite{zhai2011exploiting}, we follow their experimental settings by performing random 3-fold cross validation experiment 30 times on the other two datasets (Product and Hotel).
However, there might be slight differences on fold splitting with \newcite{zhai2011exploiting}.

In the MCCNN model, each channel has an embedding layer with a dimension of 300, randomly initialized with a uniform distribution of ${\mathcal {U}}(-0.05, 0.05)$. The filter windows within each convolution layer are 2, 3 and 4 with 256 feature maps each, where the window sizes of 2, 3 and 4 are designed to capture 2-, 3- and 4-gram features respectively. The global max pooling over the whole text sequence is adopted \cite{collobert2011natural,kim2014convolutional}. We also adopted \textit{dropout} \cite{srivastava2014dropout} with the rate of 0.6 to regularize the neural network, which was trained using the stochastic gradient descent method with the \textit{adadelta} update rule \cite{zeiler2012adadelta,kim2014convolutional}.
These hyperparameters are chosen via cross validation on the CR dataset and kept the same for all the other datasets, except for the mini-batch sizes.
In consideration of dataset size, the mini-batch sizes are set to 128 for the two large datasets CR and AD, and 16 for the two small datasets Product and Hotel.

\subsection{Results and Analysis}
\label{sec:results}

Table~\ref{tbl:cv-results} presents the comparative results among MCCNN, fastText \cite{joulin2017bag} and SVM \cite{zhai2011exploiting} with different representations on the four datasets, where \textit{Word*} means the word representation is initialized with a pre-trained Chinese word embeddings\footnote{https://fasttext.cc/docs/en/crawl-vectors.html} and the best results are underlined.
We use \texttt{fastText} as a strong baseline, which obtains state-of-the-art performance on various text classification tasks \cite{joulin2017bag,bojanowski2017enriching}. In our experiments, the settings of \texttt{fastText} are \textit{wordNgrams = 4}, \textit{epoch = 50} and \textit{minCount = 5} for all the experiments on pinyin, character and word.

\begin{table*}[htb]
\centering
	\caption{Performance comparison among fastText, MCCNN and SVM with different representations.}
	\label{tbl:cv-results}
\resizebox{\linewidth}{!}{%
\begin{tabular}{c|l|cc|cc|cc|cc}
\hline
\multirow{2}{*}{\textbf{Model}} & \multirow{2}{*}{\textbf{Representation}} & \multicolumn{2}{c|}{\textbf{CR}} & \multicolumn{2}{c|}{\textbf{AD}} & \multicolumn{2}{c}{\textbf{Product}} & \multicolumn{2}{c}{\textbf{Hotel}} \\ \cline{3-10} 
 &  & $F_1$ & Accuracy & $F_1$ & Accuracy & $F_1$ & Accuracy & $F_1$ & Accuracy \\ \hline \hline
  \multirow{6}{*}{\makecell{SVM \\ (Zhai et al., 2011)}} & \textbf{Character Unigram} & - & - & - & - & 0.9100 & 0.9100 & 0.8750 & 0.8750 \\
 & \textbf{Character Bigram} & - & - & - & - & 0.9220 & 0.9220 & \textbf{0.9190} & \textbf{0.9190} \\
 & \textbf{Character Trigram} & - & - & - & - & 0.8650 & 0.8650 & 0.9135 & 0.9130 \\
 & \textbf{Word Unigram} & - & - & - & - & \textbf{0.9280} & \textbf{0.9280} & 0.8995 & 0.8990 \\
 & \textbf{Word Bigram} & - & - & - & - & 0.8680 & 0.8680 & 0.9125 & 0.9120 \\
 & \textbf{Word Trigram} & - & - & - & - & 0.7135 & 0.7230 & 0.8775 & 0.8780 \\ \hline
\multirow{4}{*}{\makecell{fastText \\ (Joulin et al., 2017)}} & \textbf{Pinyin} & 0.8108 & 0.8091 & 0.8135 & 0.8131 & 0.9148 & 0.9148 & 0.9171 & 0.9171 \\
 & \textbf{Character} & \textbf{0.8143} & \textbf{0.8128} & 0.8131 & 0.8128 & \textbf{0.9261} & \textbf{0.9261} & \textbf{\underline{0.9218}} & \textbf{\underline{0.9218}} \\
 & \textbf{Word} & 0.8099 & 0.8076 & \textbf{0.8161} & \textbf{0.8157} & 0.9116 & 0.9116 & 0.9090 & 0.9089 \\ 
 & \textbf{Word*} & 0.8120 & 0.8100 & \textbf{\underline{0.8181}} & \textbf{\underline{0.8178}} & 0.9184 & 0.9184 & 0.9136 & 0.9136 \\ \hline
\multirow{11}{*}{\makecell{MCCNN \\ (This Paper)}} & \textbf{Pinyin} & 0.8015 & 0.8034 & 0.7816 & 0.7811 & 0.8984 & 0.8986 & \textbf{0.9097} & \textbf{0.9097} \\
 & \textbf{Character} & \textbf{0.8087} & \textbf{0.8111} & 0.7852 & 0.7852 & \textbf{0.9026} & \textbf{0.9028} & 0.9075 & 0.9076 \\
 & \textbf{Word} & 0.8053 & 0.8070 & \textbf{0.7898} & \textbf{0.7900} & 0.8502 & 0.8518 & 0.9039 & 0.9040 \\ \cline{2-10} 
 & \textbf{Pinyin + Character} & 0.8104 & 0.8121 & 0.7894 & 0.7890 & \textbf{0.9043} & \textbf{0.9045} & \textbf{0.9107} & \textbf{0.9108} \\
 & \textbf{Pinyin + Word} & 0.8115 & 0.8130 & 0.7906 & 0.7905 & 0.8773 & 0.8785 & 0.9093 & 0.9093 \\
 & \textbf{Character + Word} & 0.8136 & 0.8147 & 0.7925 & 0.7923 & 0.8891 & 0.8897 & 0.9100 & 0.9101 \\
 & \textbf{Pinyin + Character + Word} & \textbf{0.8140} & \textbf{0.8162} & \textbf{0.7954} & \textbf{0.7955} & 0.8915 & 0.8922 & 0.9076 & 0.9076 \\ \cline{2-10} 
 & \textbf{Word*} & 0.8125 & 0.8143 & 0.7956 & 0.7957 & 0.9268 & 0.9270 & 0.9197 & 0.9197 \\
 & \textbf{Pinyin + Word*} & 0.8182 & 0.8201 & \textbf{0.7989} & \textbf{0.7989} & \textbf{\underline{0.9302}} & \textbf{\underline{0.9304}} & \textbf{0.9208} & \textbf{0.9208} \\
 & \textbf{Character + Word*} & 0.8185 & 0.8211 & 0.7981 & 0.7981 & 0.9283 & 0.9285 & 0.9199 & 0.9199 \\
 & \textbf{Pinyin + Character + Word*} & \textbf{\underline{0.8200}} & \textbf{\underline{0.8216}} & 0.7979 & 0.7979 & 0.9287 & 0.9288 & 0.9201 & 0.9202 \\ \hline
\end{tabular}%
}
\end{table*}
    
\subsubsection{Comparison among Word, Character and Pinyin}
\label{sec:exp-channels}

We compare the word, character and pinyin repsentations from three perspectives:
\begin{enumerate*}[(i)]
\item token vocabulary,
\item semantic loss, and 
\item classification performance.
\end{enumerate*}

\begin{figure*}[htb]%
    \centering
    
    \subfigure[Vocabulary Size]{%
    \label{fig:voc}%
    \begin{tikzpicture}[scale=0.66]
    \begin{axis}[
    width  = 0.6\linewidth,
    height = 8cm,
    ylabel = Vocabulary Size,
    enlarge x limits=0.2,
    ymode=log,
    ybar,
    xtick=data,
    symbolic x coords={CR, AD, Product, Hotel},
    grid=major,
    xmajorgrids=false
    ]
  
    \addplot coordinates {(CR, 294122) (AD,155862) (Product,4010) (Hotel,14924)};
    \addplot coordinates {(CR, 6469) (AD,5212) (Product,1485) (Hotel,2761)};
    \addplot coordinates {(CR, 1325) (AD,1304) (Product,825) (Hotel,1100)};
    
    \legend{Word, Character, Pinyin}
    \end{axis}
    \end{tikzpicture}}
    \qquad
    \subfigure[OOV Rate]{%
  \begin{tikzpicture}[scale=0.66]
\begin{axis}[
    width  = 0.6\linewidth,
    height = 8cm,
    ylabel = OOV Rate,
    enlarge x limits=0.2,
    ybar,
    xtick=data,
    symbolic x coords={CR, AD, Product, Hotel},
    grid=major,
    xmajorgrids=false,
    ]

    \addplot coordinates {(CR, 0.1448) (AD, 0.1173) (Product, 0.4019) (Hotel, 0.2850)};
    \addplot coordinates {(CR, 0.0136) (AD, 0.0106) (Product, 0.1296) (Hotel, 0.0721)};
    \addplot coordinates {(CR, 0.0015) (AD, 0.0014) (Product, 0.0616) (Hotel, 0.0226)};
    
    \legend{Word, Character, Pinyin}
\end{axis}
\end{tikzpicture}}
    \vspace{-1em}
    \caption{Comparison among \textit{word}, \textit{character} and \textit{pinyin} on \textit{Vocabulary Size} and \textit{OOV Rate}.}
    \label{fig:voc-size}
\end{figure*}
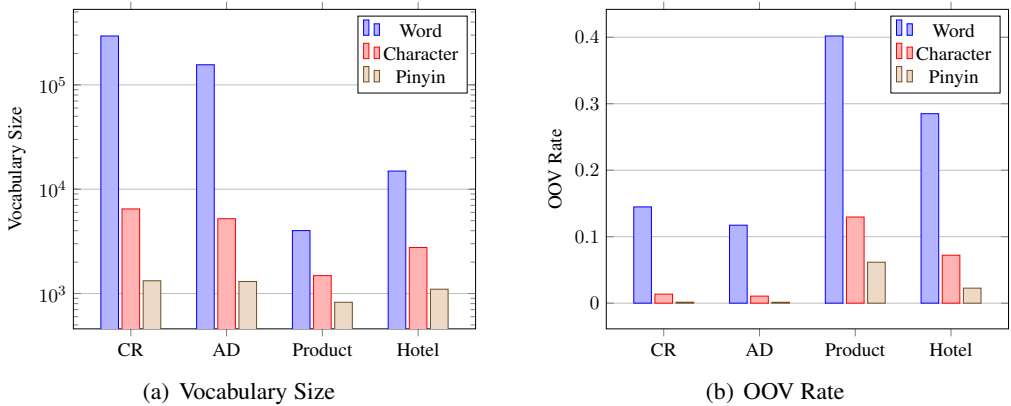

{\textbf{Token Vocabulary}}
Different representations lead to different token vocabularies. A word, character or pinyin is considered as a token. Figure~\ref{fig:voc-size} (a) shows the vocabulary size of each representation for the four datasets. The total number of words can be as large as hundreds of thousand, while the total number of characters is a few thousand. Pinyin has the smallest vocabulary size, less than two thousand on all the datasets. In terms of pre-processing effort, the character representation requires the least effort by just splitting a text into individual characters, whereas both of the other two representations need to perform Chinese word segmentation first. Another interesting comparison is on OOV rate in Figure~\ref{fig:voc-size} (b), where the word representation has the largest OOV rate particularly on small datasets like Product and Hotel. The pinyin representation has the smallest OOV rate, while the character representation has less than 0.1 OOV rate on the four datasets except the Product dataset. Note that OOV rate is a useful reference when choosing a representation since a model can not perform well on an unseen dataset with high OOV rate.

\begin{figure*}[htb]%
    \centering
    \subfigure[Word]{%
    \label{fig:word}%
    \includegraphics[width=0.3\linewidth]{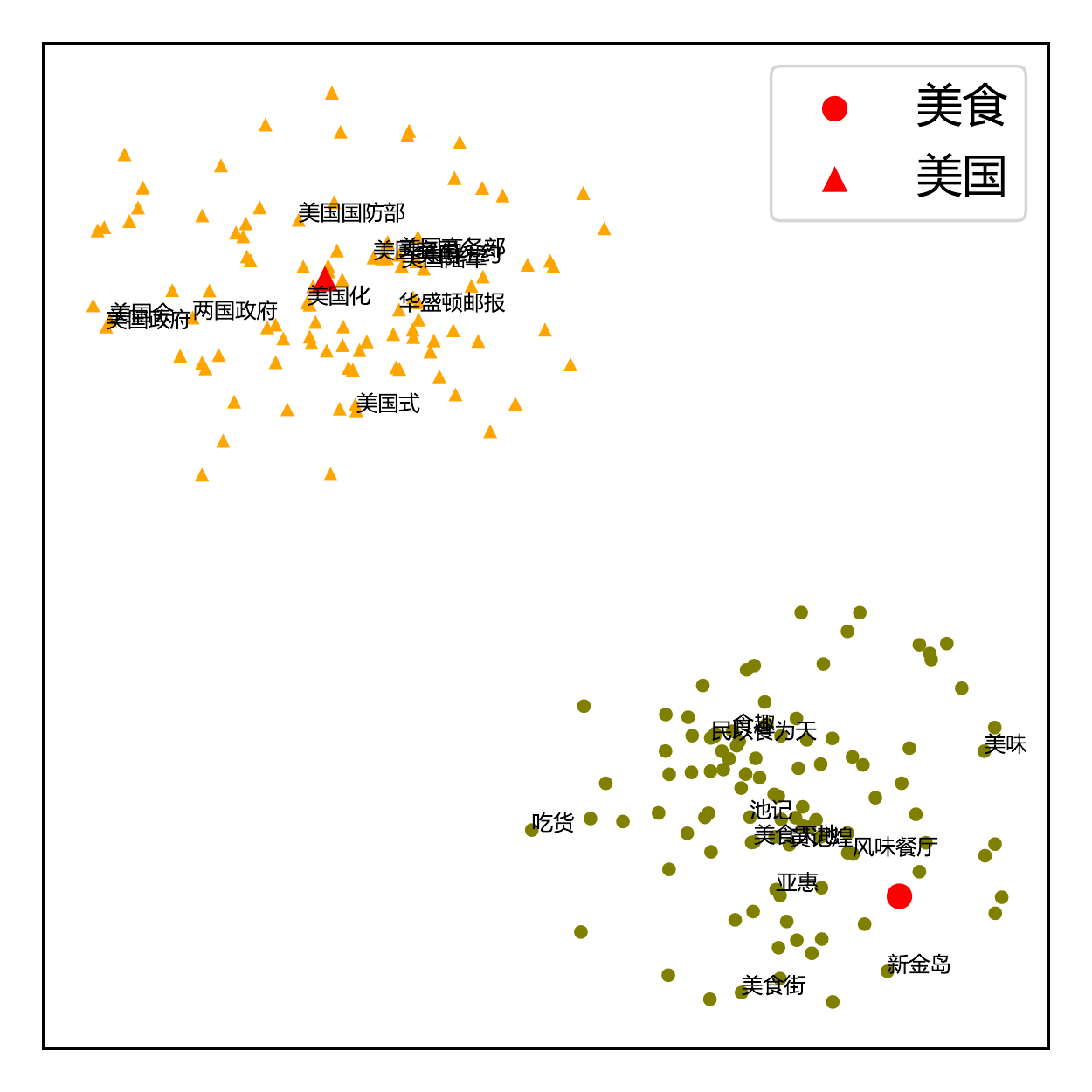}}%
    \qquad
    \subfigure[Character]{%
    \label{fig:char}%
    \includegraphics[width=0.3\linewidth]{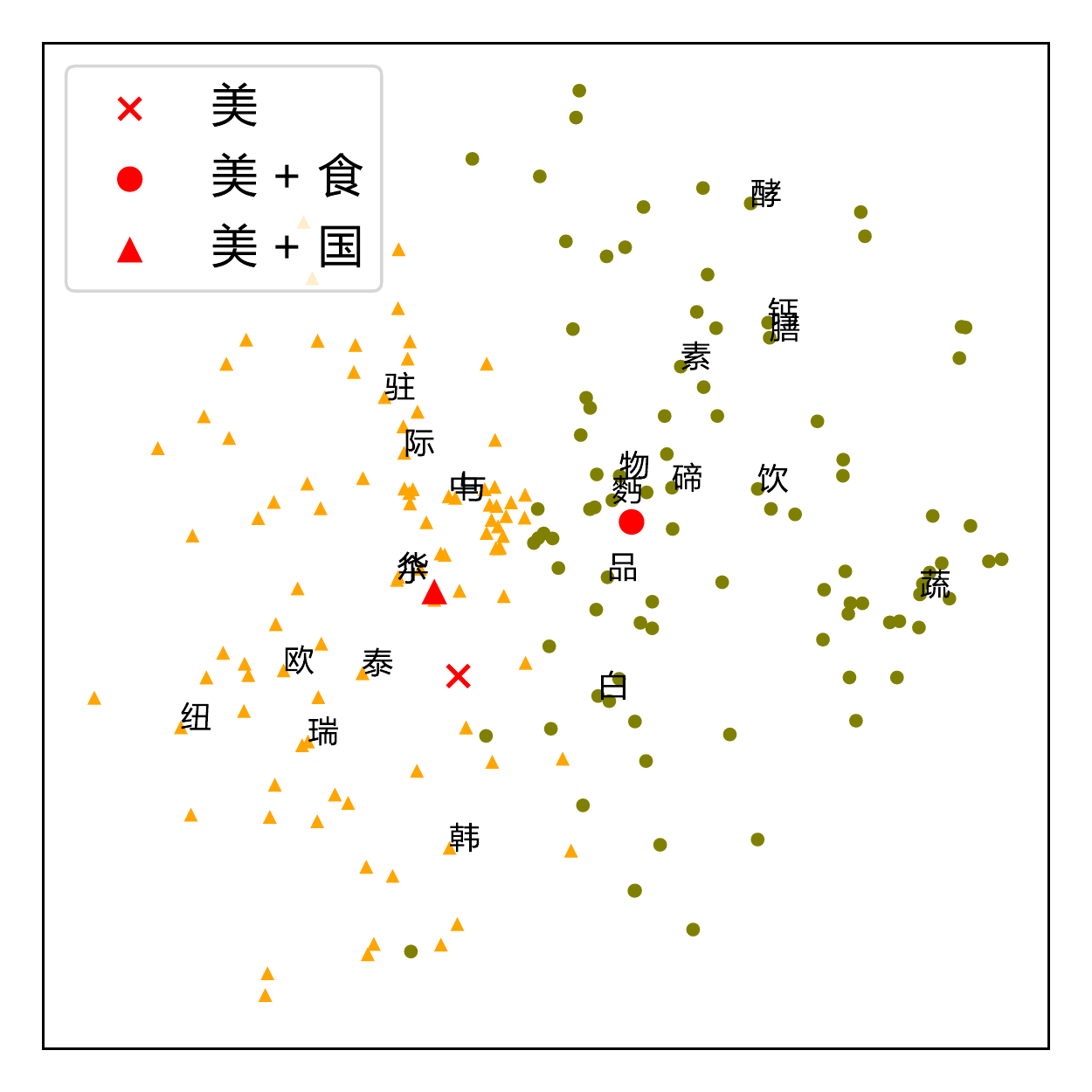}}%
    \qquad
    \subfigure[Pinyin]{%
    \label{fig:pinyin}%
    \includegraphics[width=0.3\linewidth]{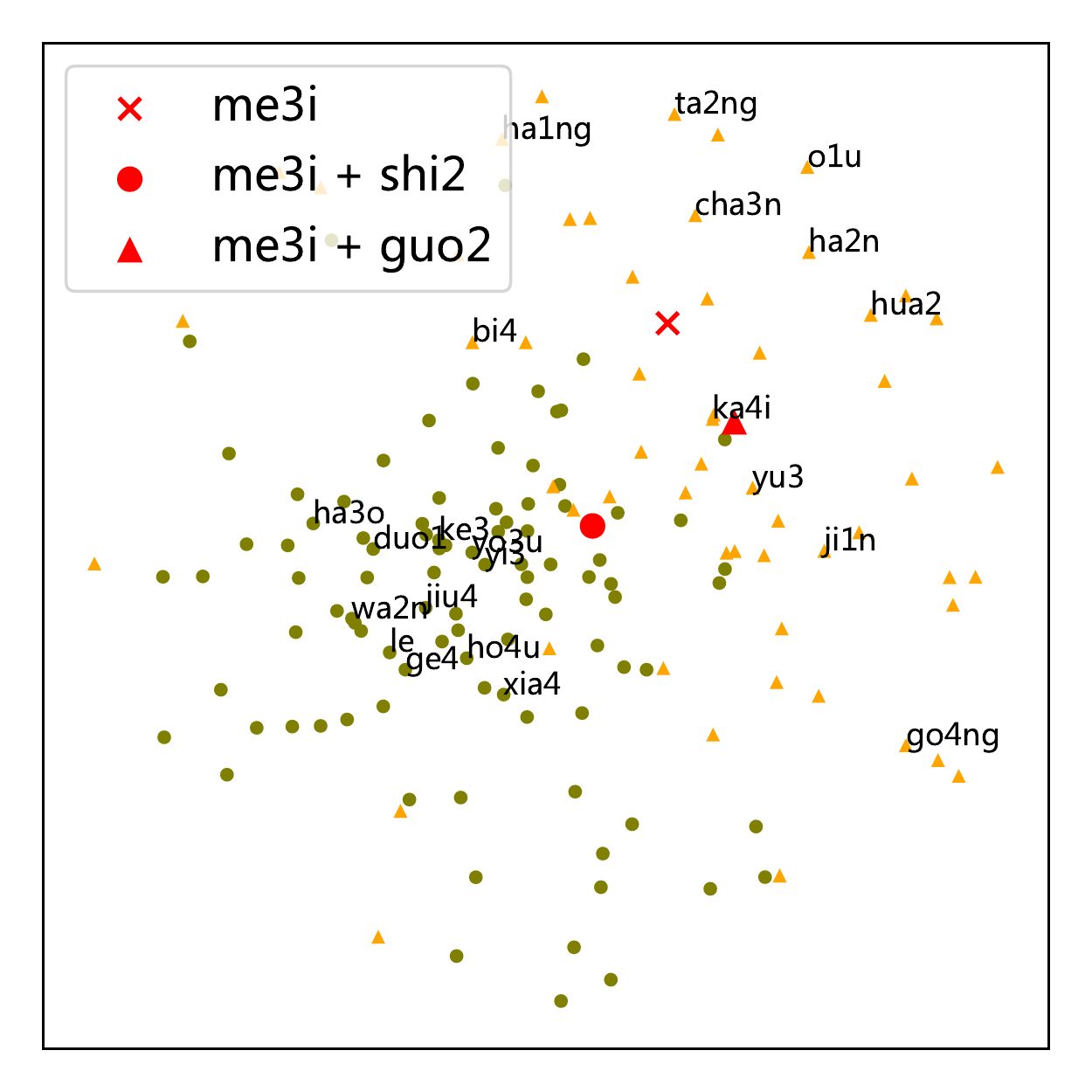}}%
    \caption{Embeddings of example words, characters and pinyins visualized using t-SNE.} 
    \label{fig:embed-visual}
\end{figure*}

{\textbf{Semantic Loss}}
Semantic loss happens when the meaning of a text is partially or completely lost under a particular representation.
We compare the semantic loss caused by the word, character and pinyin representations from the embedding perspective, by learning their embeddings using \texttt{fastText} with all the data from CR and AD.  Figure~\ref{fig:embed-visual} visualizes the resulting embeddings of example words. There are two groups of words in Figure~\ref{fig:word}, where the first group is related to \begin{CJK*}{UTF8}{gbsn}美食\end{CJK*} (tasty food) in yellow, and the second group is related to \begin{CJK*}{UTF8}{gbsn}美国\end{CJK*} (United States) in green. Figure~\ref{fig:char} also presents two groups of semantically similar characters obtained by averaging the embeddings of the corresponding characters (e.g., \begin{CJK*}{UTF8}{gbsn}美+食, 美+国\end{CJK*}) and finding out their most similar characters. However, the pinyins in each group of Figure~\ref{fig:pinyin} are not necessarily semantically similar as multiple characters may have the same pinyin. Figure~\ref{fig:embed-visual} clearly shows the loss of semantics from word to character and pinyin, which may indicate that character representation appears to be a good trade-off between decreasing \textit{vocabulary size / OOV rate} while retaining semantics. 

{\textbf{Classification Performance}}
By comparing the classification performance of word, character and pinyin in Table~\ref{tbl:cv-results}, we have the following observations: 
\begin{enumerate}[(1), itemsep=0.2ex]

    \item For both MCCNN and \texttt{fastText}, the word representation wins on AD which has the smallest OOV rate (11.73\%), while the character representation wins on the other three datasets.
    
    \item The character representation and the pinyin representation obtain similar performance on all datasets by both MCCNN and \texttt{fastText}. Most of the accuracy differences are not significant ($p>0.05$), e.g., on Product and Hotel.
    
    \item Generally, the pinyin representation alone obtains competitive performance compared with character and word, where the maximum accuracy difference is less than 1\%. This is surprising, since the pinyin representation loses much information about the original text, e.g., boundary information about words, different characters sharing the same pinyins.
    
\end{enumerate}

\subsubsection{Comparison between MCCNN and fastText}
\label{sec:mccnn-fasttext}

MCCNN with \textit{Pinyin + Character + Word*} outperforms \texttt{fastText} with \textit{Character} significantly on CR (141 words on average) with $p<9.1\mathrm{e}{-7}$. However, \texttt{fastText} outperforms MCCNN on AD (45 words on average) with $p<5.4\mathrm{e}{-8}$.
MCCNN and \texttt{fastText} obtain close performance on the other two small datasets: MCCNN outperforms \texttt{fastText} slightly on Product (34 words on average) while \texttt{fastText} wins on Hotel (69 words on average).

Generally, \texttt{fastText} shows better performance than MCCNN on the datasets with short documents while MCCNN performs significantly better than \texttt{fastText} on the dataset (i.e., CR) with very long documents. We attribute that \texttt{fastText}, which essentially \textit{averages} all ngram features of a text for prediction, may become ineffective in capturing semantic compositionality \cite{socher2013recursive} as document lengths increase. On the other hand, MCCNN can effectively extract local features with feature maps of the convolution layers even from very long documents.

Note that \texttt{fastText} does not support multi-channel input (e.g., Word + Character + Pinyin), which prevents us from comparing MCCNN with \texttt{fastText} in the multi-channel setting. 
To support multi-channel input in \texttt{fastText}, a possible extension is to add additional separate input layers for multiple channels, and add a \textit{concatenation} layer after the \textit{average} layer in \texttt{fastText}. We leave this extension as future work.

\subsubsection{Effect of Representation Combination}

Overall, Table~\ref{tbl:cv-results} shows that the three representations complement each other, since using them in combination usually achieves better performance than using a single representation.
For example, \textit{Pinyin + Character + Word} outperforms the best single representation of \textit{Character} on CR with $p=0.0031$ and outperforms the \textit{Word} representation on AD with $p=0.0001$.
However, the performance improvement is not always significant. For example, \textit{Pinyin + Character} improves \textit{Pinyin} with $p=0.0094$ on Product but not on Hotel with $p=0.3386$. Adding \textit{Pinyin} or \textit{Character} or both over \textit{Word} does not improve the performance significantly either ($p > 0.05$).

\subsubsection{Effect of Pre-trained Embeddings}

It is a common practice to adopt pre-trained embeddings in NLP tasks. The performance improvement by the pre-trained embeddings is also observed in Table~\ref{tbl:cv-results} for both MCCNN and \texttt{fastText} on all the datasets.
For example, initializing the word channel with the pre-trained word embeddings (\textit{Pinyin + Character + Word*}) outperforms random embeddings (\textit{Pinyin + Character + Word}) significantly with $p<0.002$ on CR.

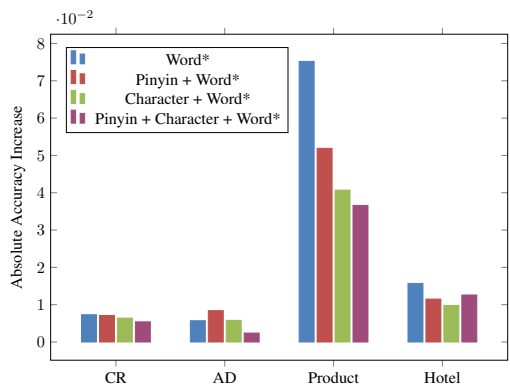
\begin{figure}[htb]
\centering
\begin{tikzpicture}[scale=0.56]
\begin{axis}[
    width  = 1.6\linewidth,
    height = 9.3cm,
    ylabel = Absolute Accuracy Increase,
    enlarge x limits=0.2,
    legend style={
        at={(0.52,0.96)},
        column sep=1ex
    },
    ybar,
    xtick=data,
    symbolic x coords={CR, AD, Product, Hotel},
    cycle list/Dark2, every axis plot/.append style={fill}
    ]
    
    \addplot[style={bblue,fill=bblue,mark=none}] coordinates {(CR, 0.0073) (AD, 0.0057) (Product, 0.0752) (Hotel, 0.0157)};
    \addplot[style={rred,fill=rred,mark=none}] coordinates {(CR, 0.0071) (AD, 0.0084) (Product, 0.0519) (Hotel, 0.0115)};
    \addplot[style={ggreen,fill=ggreen,mark=none}] coordinates {(CR, 0.0064) (AD, 0.0058) (Product, 0.0407) (Hotel, 0.0098)};
    \addplot[style={ppurple,fill=ppurple,mark=none}] coordinates {(CR, 0.0054) (AD, 0.0024) (Product, 0.0366) (Hotel, 0.0126)};
    
    \legend{Word*, Pinyin + Word*, Character + Word*, Pinyin + Character + Word*}
\end{axis}
\end{tikzpicture}
\caption{Absolute accuracy increase by adopting pre-trained word embeddings.}
\label{fig:pre-embed}
\end{figure}

Figure~\ref{fig:pre-embed} shows the absolute accuracy increase by adopting pre-trained word embeddings in the four settings of MCCNN.
We observe that the pre-trained word embeddings are particularly important for the small datasets with high OOV rate (e.g., Product).
However, the performance improvement is not always significant on the large dataset with relatively lower OOV rate. For example, the improvement on AD by \textit{Pinyin + Character + Word*} is not significant with $p>0.05$.

\subsubsection{Effect of Filter Window Size}

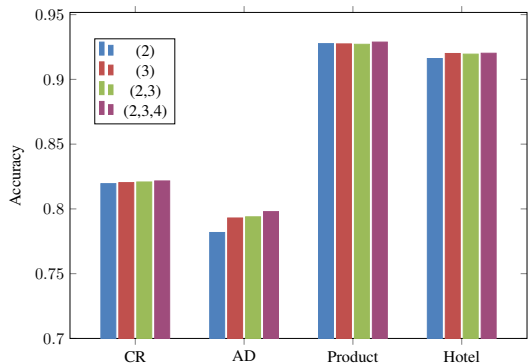
\begin{figure}[htb]
\centering
\begin{tikzpicture}[scale=0.56]
\begin{axis}[
    width  = 1.6\linewidth,
    height = 9.3cm,
    ylabel = Accuracy,
    enlarge x limits=0.2,
    legend style={
        at={(0.23,0.92)},
        column sep=1ex
    },
    ymin=0.700,
    ybar,
    xtick=data,
    symbolic x coords={CR, AD, Product, Hotel},
    cycle list/Dark2, every axis plot/.append style={fill}
    ]
    
    \addplot[style={bblue,fill=bblue,mark=none}] coordinates {(CR, 0.8195) (AD, 0.7818) (Product, 0.9276) (Hotel, 0.9161)};
    \addplot[style={rred,fill=rred,mark=none}] coordinates {(CR, 0.8203) (AD, 0.7930) (Product, 0.9274) (Hotel, 0.9200)};
    \addplot[style={ggreen,fill=ggreen,mark=none}] coordinates {(CR, 0.8208) (AD, 0.7939) (Product, 0.9271) (Hotel, 0.9195)};
    \addplot[style={ppurple,fill=ppurple,mark=none}] coordinates {(CR, 0.8216) (AD, 0.7979) (Product, 0.9288) (Hotel, 0.9202)};
    
    \legend{(2), (3), (2,3), (2,3,4)}
\end{axis}
\end{tikzpicture}
\caption{Accuracy comparison among different filter window sizes.}
\label{fig:filter-size}
\end{figure}

We vary the size of filter windows in MCCNN fed with \textit{Pinyin + Character + Word*} to investigate its effect on classification performance. As shown in Figure~\ref{fig:filter-size}, the setting of (2,3,4) for filter windows gives the best accuracy on the four datasets. Interestingly, the accuracy differences among various window sizes are relatively small. For example, the standard deviations of accuracy are 0.0007 and 0.0019 for Product and Hotel respectively. 
Note that the stable performance of CNN with respect to different filter window sizes is also observed by \newcite{zhang2017sensitivity}.

\section{Conclusion}

In this paper, we exploited effective Chinese representations for sentiment analysis. To facilitate the investigation, we proposed a multi-channel convolutional neural network (MCCNN), where each channel corresponds to one particular representation of a Chinese text, namely \textit{word}, \textit{character} and \textit{pinyin}.
Notably, the character representation is efficient in terms of small vocabulary size and requires the least pre-processing effort without the need for Chinese word segmentation.
Experimental results show that the word representation obtains the best performance on the dataset with low OOV rate while the character representation performs the best on the datasets with high OOV rate. Although losing much information, the pinyin representation achieves surprisingly good performance comparable with either word or character. Using these representations in combination generally achieves better performance, indicating that they complement each other.
Initializing the word channel with pre-trained word embeddings improves MCCNN further.
Besides, MCCNN outperforms SVM based on conventional ngram features and is competitive with the state-of-the-art \texttt{fastText} model.




\bibliography{references}
\bibliographystyle{acl_natbib_nourl}

\end{document}